\documentclass[conference]{IEEEtran}
\IEEEoverridecommandlockouts
\usepackage[sort&compress,numbers]{natbib}
\usepackage{csquotes}
\usepackage{amsmath,amssymb,amsfonts}
\usepackage{algorithmic}
\usepackage{graphicx}
\usepackage{textcomp}
\usepackage{tikz} 
\usepackage{xcolor}
\definecolor{promptboxcolor}{RGB}{193, 255, 193} 
\usetikzlibrary{positioning} 
\usepackage{tabularx}
\usepackage{booktabs}
\usepackage{amsmath, amsthm, amsfonts}
\usepackage[table]{xcolor}
\newcommand{\best}[1]{\cellcolor{green!20}\textbf{#1}} 
\newcommand{\worst}[1]{\cellcolor{red!20}#1}
\usepackage{mathtools}

\usepackage{multirow}

\theoremstyle{definition}
\newtheorem{definition}{Definition}
\usepackage{tabularx, booktabs}

\usepackage{amsmath, amsfonts, amssymb, amsthm}
\usepackage{url}

\def\BibTeX{{\rm B\kern-.05em{\sc i\kern-.025em b}\kern-.08em
    T\kern-.1667em\lower.7ex\hbox{E}\kern-.125emX}}

\begin{document}


\title{TopoQuant: Topological Framework to Characterize Behavioral Shifts in Quantized Large Language Models}


\author{
\IEEEauthorblockN{
Baha Rababah$^{1,2}$,
Shahzeb Qamar$^3$,
Lorenz Sparrenberg$^3$,
Rafet Sifa$^3$\\
Murat Kantarcioglu$^4$,
Carson K. Leung$^1$,
Cuneyt Gurcan Akcora$^5$
}
\IEEEauthorblockA{
\textit{$^1$ Department of Computer Science, University of Manitoba},
Winnipeg, MB, Canada\\
\textit{$^2$ Applied Computer Education Department, Red River College Polytechnic},
Winnipeg, MB, Canada\\
\textit{$^3$ Bonn-Aachen International Center for IT (b-it), University of Bonn},
Bonn, Germany\\
\textit{$^4$ Department of Computer Science, Virginia Tech University},
Blacksburg, VA, USA\\
\textit{$^5$ AI Initiative, University of Central Florida},
Orlando, FL, USA\\
rababahb@myumanitoba.ca, 
shahzeb.qamar@iais.fraunhofer.de, 
lsparren@uni-bonn.de,\\
Rafet.Sifa@bit.uni-bonn.de, 
muratk@vt.edu, 
Carson.Leung@UManitoba.ca,
cuneyt.akcora@ucf.edu}}

\maketitle

\begin{abstract}

Post-training quantization (PTQ) reduces the computational cost of large language models (LLMs), but it can introduce internal structural and behavioral changes that are not captured by traditional metrics such as accuracy and perplexity. We present \textbf{TopoQuant}, a topological framework that uses topological data analysis (TDA) to quantify structural drift in self-attention weights and relate it to behavioral changes in quantized LLMs. Across four LLM families and multiple quantization levels, we show that topological drift remains stable at moderate compression levels but increases sharply below 4-bit precision, closely tracking declines in attention agreement. Our results reveal a strong relationship between topological drift and attention agreement, indicating that structural changes in model parameters closely reflect behavioral degradation under quantization. We further demonstrate that topological features can accurately predict attention consistency using lightweight machine learning models, achieving $R^2$ scores of up to 0.85. Predicting attention agreement enables efficient assessment of behavioral consistency in quantized LLMs, allowing quantization-induced degradation to be detected without running expensive downstream tasks. The results indicate that topological drift provides an efficient early-warning signal for quantization-induced degradation and can help identify safe compression limits without extensive downstream evaluation.
\end{abstract}

\begin{IEEEkeywords}
Post-Training Quantization; Large Language Models; Topological Data Analysis; Persistent Homology; Attention Mechanisms; Behavioral Evaluation; Machine Learning
\end{IEEEkeywords}

\section{Introduction}

Post-training quantization (PTQ), which reduces the precision of pre-trained large language model (LLM) parameters, has emerged as an effective approach for lowering memory consumption, inference latency, and computational costs during deployment~\cite{Zhu2024survey,rababah2026post}. However, quantization often introduces performance degradation and behavioral drift~\cite{duttaKKR24accuracy}. Traditional evaluation metrics, such as accuracy, loss, and perplexity, provide only limited insights into how and why quantization affects the model behavior~\cite{duttaKKR24accuracy,rababah2026illusion}.

Understanding these effects requires examining the internal structure of the model, rather than relying solely on downstream performance. Topological data analysis (TDA) offers a promising framework for this purpose by characterizing the shape, connectivity, and structural organization of high-dimensional data~\cite{edelsbrunner2013persistent,carlsson2004persistence}. Consequently, topological features may reveal quantization-induced changes that remain hidden from conventional statistical metrics.


Attention mechanisms play a central role in transformer-based LLMs by determining which tokens influence model predictions~\cite{VaswaniSPUJGKP17}. Prior studies have shown that attention behavior is sensitive to quantization artifacts and closely linked to model behavior~\cite{BondarenkoNB23,ChengDQW25,kadem2026}. Quantization-induced perturbations to the Query, Key, and Value projections can alter attention allocation patterns, even when the overall task accuracy remains stable. Therefore, measuring the attention agreement between full-precision and quantized models provides a sensitive indicator of behavioral drift that may not be detected using aggregate performance metrics alone.

Recent parameter-based quantization evaluation frameworks have attempted to analyze quantization effects directly from model parameters~\cite{duttaKKR24accuracy,rababah2026illusion}. However, several limitations remain. First, high accuracy can be misleading because quantized models may produce substantially different predictions while maintaining similar aggregate performance. Second, existing approaches primarily rely on aggregate statistical or coordinate-based representations that quantify numerical drift but fail to capture the underlying geometric and structural organization of model parameters. Third, current frameworks~\cite{jin2024Comprehensive, Zhao2025Benchmarking}  lack effective mechanisms for early detection of quantization-induced degradation. Finally, most approaches aggregate Query ($Q$), Key ($K$), and Value ($V$) projections into a single global measure~\cite{rababah2026illusion}, making it difficult to identify which attention components are most affected by quantization.


To address these limitations, we propose \textbf{TopoQuant}, a topological framework for analyzing quantization-induced changes in large language models (LLMs). The framework combines three complementary perspectives: behavioral drift analysis, attention consistency analysis, and topological drift analysis. Attention consistency is quantified using Top-1 agreement and Top-3 Intersection-over-Union (IoU), whereas structural drift is characterized using TDA-based representations. By relating topological changes to attention behavior, TopoQuant provides an interpretable view of how quantization alters the model behavior and internal organization.

TopoQuant analyzes Query ($Q$), Key ($K$), and Value ($V$) projection matrices independently, enabling targeted diagnosis of quantization-sensitive components. We focus on these projections because they directly govern attention formation and token interactions, whereas the Output ($O$) projection primarily transforms attention outputs after the attention distribution has already been computed. This design allows the framework to identify structural changes before they become apparent in conventional downstream evaluation metrics.

This study investigates LLMs quantized from 8-bit to 2-bit precision and explores whether topological representations can characterize and predict quantization-induced behavioral changes. Our main contributions are:

 \begin{itemize} 
 \item We propose \textbf{TopoQuant}, a topology-aware framework for quantifying quantization-induced structural drift in LLM attention weights using persistence diagrams. 
 
 \item We introduce a multi-metric topological analysis pipeline that measures structural changes between base and quantized models across attention projections, blocks, and quantization levels.
 
 \item We link topological drift to attention behavior through Top-1 agreement and Top-3 IoU-based attention consistency analysis. 
 
 \item We show that topological drift can identify quantization degradation and determine effective quantization stopping points without exhaustive model evaluation. 
 
 \item We develop lightweight machine learning models that predict attention agreement between the base and quantized model directly from topological drift features and LLM characteristics.
 
 \item We evaluate TopoQuant across multiple LLM families and quantization schemes, showing its effectiveness in characterizing and predicting changes in quantization behavior. \end{itemize}

\section{Related Work}
\label{sec:Related Work}


\textbf{PTQ methods and evaluation frameworks.} PTQ is widely used to reduce the memory footprint and computational cost of LLMs while avoiding costly retraining. Representative methods include GPTQ~\cite{frantar2022gptq}, OmniQuant~\cite{Shao2024OmniQuant} and SliM-LLM~\cite{huang2024slim}. Several studies have proposed evaluation frameworks to assess quantized models across performance, efficiency, robustness, and alignment dimensions~\cite{jin2024Comprehensive,Kurtic2025give,Zhao2025Benchmarking}. While these studies provide valuable benchmarks, they primarily focus on task-level performance and do not investigate the structural changes introduced by quantization or their relationship with behavioral drift.

\textbf{Limitations of perplexity and accuracy.} Several studies have shown that conventional metrics such as accuracy and perplexity may fail to capture important behavioral differences between full-precision and quantized LLMs~\cite{kumar2024compressing,Deiseroth2024divergent}. Alternative measures, including KL divergence, prediction-flip rates \cite{duttaKKR24accuracy}, correctness agreement \cite{rababah2026illusion}, Jensen--Shannon divergence, and total variation distance, have been proposed to quantify behavioral and distributional shifts~\cite{duttaKKR24accuracy,Nikolic2026displacment,rababah2026illusion,qamar2026accuracy}. Although these approaches reveal behavioral changes that aggregate metrics often miss, they do not explicitly capture the structural transformations occurring within the model parameters.

\textbf{Structural analysis using TDA.} TDA has been increasingly applied to deep learning to analyze neural network complexity, representation dynamics, robustness, and interpretability~\cite{hofer2017deep,rieck2018neural,carriere2020perslay,gardinazzi2024persistent,chauhan2022bertops}. Recently, TDA has been explored for evaluating quantized neural networks~\cite{ogio2025tda}. However, its use for studying quantization-induced behavioral drift and attention consistency in LLMs remains largely unexamined.

Despite these advances, existing approaches primarily evaluate quantization using task performance and output distrmmulti-scaleor conventional statistical properties. These perspectives provide limited insight into the multiscale structural changes occurring within attention weight matrices and how these changes relate to behavioral consistency.

\section{Notation and Core Definitions}
\label{sec:Notation_and_Definitions}

In this section, we formalize PTQ as an operator on transformer parameters and define two metric spaces: (1) structural drift in the weight space and (2) decision-level consistency in the output space.

\subsection{Notation and Model Formalism}
\label{subsec:Notation_and_Model_Formalism}
Let $\mathcal{V}$ be a vocabulary and $\mathcal{X} \subseteq \mathcal{V}^\star$ be the set of token sequences. We consider a decoder-only transformer with $N$ blocks and hidden dimension $d$, parameterized by $\theta \in \mathbb{R}^P$, where $P$ is the total parameter count. For an input prompt $x \in \mathcal{X}$ and target sequence $y = (y_1, \dots, y_T) \in \mathcal{V}^\star$, the model induces next-token conditional probability distributions:
\begin{equation}
p_\theta(y \mid x) = \prod_{t=1}^T p_\theta(y_t \mid x, y_{<t}).
\end{equation}

For each transformer block index $i \in [N] := \{1, \dots, N\}$, let $W_i^Q, W_i^K, W_i^V \in \mathbb{R}^{d \times d}$ denote the self-attention projection matrices. We collect all attention parameters into the set:
\begin{equation}
\mathcal{W}_\theta:= \left\{ W_i^L: i \in [N], \, L \in \{Q, K, V\} \right\}.
\end{equation}

\subsection{Post-Training Quantization (PTQ) Operators}
\label{subsec:PTQ_Operators}

We define a finite set of quantization configurations $\mathcal{C}$, where each $c \in \mathcal{C}$ specifies a bit width $b(c) \in \{2, 3, 4, 5, 6, 8\}$ and a quantization scheme (e.g., K-quantization). Let $\mathcal{Q}_c$ and $\mathcal{D}_c$ represent the quantization and dequantization mapping functions. We model post-training quantization as an operator $T_c: \mathbb{R}^P \to \mathbb{R}^P$ acting on parameter space:
\begin{equation}
T_c(\theta) := \mathcal{D}_c(\mathcal{Q}_c(\theta)) \in \mathbb{R}^P.
\end{equation}

The resulting quantized-dequantized model parameterization is denoted as $T_c(\theta)$, and its corresponding parameter matrix for the attention projection layer $L \in \{Q, K, V\}$ at block $i \in [N]$ is defined as $\widehat{W}_i^{L,(c)} := W_i^L(T_c(\theta))$. 

For a given block $i$ and projection layer $L$, the entry-wise weight error matrix is defined as:
\begin{equation}
\Delta_{W, i}^{L,(c)} := W_i^L - \widehat{W}_i^{L,(c)} .
\end{equation}
When evaluating across all parameters in the concatenated parameter vector $\theta$, the overall parameter perturbation vector is denoted as $\Delta_\theta^{(c)} := T_c(\theta) - \theta$.

\subsection{Weight-Space Topological Drift}
\label{subsec:Weight_Space_Topological_Drift}

To quantify the structural changes induced by quantization in the
attention projection weights, we considered a family of topological
distance measures evaluated independently for each transformer block
and projection type. Let $ \mathcal{M} =[ \mathrm{Betti}, \mathrm{Entropy}, \mathrm{Landscape}, \mathrm{PFD}, \mathrm{PI}, \mathrm{SWD}, \mathrm{TSWD}]$
denote the set of topological distance measures considered~\cite{a_source_that_defines_them}. For
$M \in \mathcal{M}$, let
$\delta_M(\cdot,\cdot)$ denote the corresponding distance between the
topological representations of the two weight matrices.

For transformer block $i \in [N]$ and projection type
$L \in \{Q,K,V\}$, the topological distance induced by quantization
configuration $c$ is defined as
\begin{equation}
\delta_{M,i}^{L,(c)}(\theta)
:=
\delta_M
\left(
\widehat{W}_i^{L,(c)},
W_i^L
\right).
\label{eq:block_topological_distance}
\end{equation}

The averaged topological distance for projection type $L$ is then
defined by
\begin{equation}
\overline{\delta}_{M}^{L,(c)}(\theta)
:=
\frac{1}{N}
\sum_{i=1}^{N}
\delta_{M,i}^{L,(c)}(\theta),
\label{eq:layer_averaged_distance}
\end{equation}
where $M \in \mathcal{M}$.

\subsection{Behavioral Consistency Metrics}
\label{subsec:Behavioral_Consistency}

To evaluate the model behavior on downstream task datasets $\mathcal{D} = \{(x^{(s)}, y^{(s)})\}_{s=1}^S$, let $g_\theta(x)$ denote the model's predicted output under a deterministic scoring functional.

For each instance $s \in [S]$, we define binary correctness indicators for the base model $M_0$ (parameterized by $\theta$) and quantized model $M_q$ (parameterized by $T_c(\theta)$):
\begin{align}
C_0(x^{(s)}) &= \mathbb{I}\left(g_\theta(x^{(s)}) = y^{(s)}\right), \\
C_q(x^{(s)}) &= \mathbb{I}\left(g_{T_c(\theta)}(x^{(s)}) = y^{(s)}\right).
\end{align}

Standard task-level accuracies are given by the expectations $\text{Acc}(M_0) = \mathbb{E}[C_0(x)]$ and $\text{Acc}(M_q) = \mathbb{E}[C_q(x)]$. However, aggregate accuracy completely ignores instance-level behavioral stability that is, whether $M_q$ preserves correct predictions on the exact same individual prompts relative to $M_0$.

\begin{definition}[Top-1 Exact Match Token]
\label{def:top1_exact_match}
Let $x = (x_1, \dots, x_T) \in \mathcal{X}$ be an input prompt sequence of length $T = |x|$, with query position index $q = T$. Let $A_i^{(h)} \in \mathbb{R}^{T \times T}$ denote the attention matrix at transformer block $i \in [N]$ for attention head $h \in \{1, \dots, H\}$, where $A_{i, q, t}^{(h)}$ represents the attention score assigned by the query token $x_q$ to target token $x_t$.

The head-averaged attention weight vector for the query token $x_q$ at block $i$ is defined as:
\begin{equation}
\overline{A}_{i, q}(t) \triangleq \frac{1}{H} \sum_{h=1}^H A_{i, q, t}^{(h)}, \quad \forall t \in \{1, \dots, T\}.
\end{equation}

To restrict evaluation strictly to non-special context tokens, such as markers for the beginning or end of a sequence, let $\mathcal{S} \subset \mathcal{V}$ denote the set of special vocabulary tokens, and define the masked attention vector $\widetilde{A}_{i, q}(t)$ by setting self-attention and special token scores to $-\infty$:
\begin{equation}
\widetilde{A}_{i, q}(t) \triangleq 
\begin{cases}
-\infty, & \text{if } t = q \text{ or } x_t \in \mathcal{S}, \\
\overline{A}_{i, q}(t) & \text{otherwise.}
\end{cases}
\end{equation}

The Top-1 Exact Match token index $t^\star \in \{1, \dots, T - 1\}$ is defined as the context position attaining the maximum masked head-averaged attention score:
\begin{equation}
t^\star \triangleq \arg\max_{t \in \{1, \dots, T\}} \widetilde{A}_{i, q}(t).
\end{equation}
\end{definition}

\begin{definition}[Top-3 IoU Attention Similarity]
\label{def:top3_iou_similarity}
Let $K = 3$. For a given prompt sequence $x^{(s)} \in \mathcal{D}$ at transformer block $i \in [N]$, let $\mathcal{T}_0^{(K)}(s, i) \subseteq \mathcal{V}$ denote the ordered set of top-$K$ attended non-special context tokens retrieved under the base parameterization $\theta$. Similarly, let $\mathcal{T}_q^{(K)}(s, i) \subseteq \mathcal{V}$ denote the corresponding set under the quantized parameterization $T_c(\theta)$.

Filtering out null or unassigned elements, let the sets of non-null extracted tokens be defined as:
\begin{equation}
\begin{aligned}
\mathcal{U}_0 &\triangleq \left\{ t \in \mathcal{T}_0^{(K)}(s, i) \mid t \neq \varnothing \right\}, \\
\text{and } \quad \mathcal{U}_q &\triangleq \left\{ t \in \mathcal{T}_q^{(K)}(s, i) \mid t \neq \varnothing \right\}.
\end{aligned}
\end{equation}

The Top-3 Intersection over Union ($\mathrm{IoU}$) similarity measure between the base model $M_0$ and the quantized model $M_q$ at block $i$ for instance $s$ is defined using the Jaccard index:
\begin{equation}
\mathrm{IoU}_K(M_0, M_q; s, i) \triangleq 
\begin{cases}
1.0, & \text{if } \mathcal{U}_0 =  \mathcal{U}_q= \emptyset,\\
\dfrac{|\mathcal{U}_0 \cap \mathcal{U}_q|}{|\mathcal{U}_0 \cup \mathcal{U}_q|} & \text{otherwise.}
\end{cases}
\end{equation}

The aggregate Top-3 IoU similarity across all $S$ instances in dataset $\mathcal{D}$ for block $i$ is given by the expectation:
\begin{equation}
\overline{\mathrm{IoU}}_K(M_0, M_q; i) \triangleq \frac{1}{S} \sum_{s=1}^S \mathrm{IoU}_K(M_0, M_q; s, i).
\end{equation}
\end{definition}

\section{TopoQuant Design}
\label{sec:TopoQuant_Design}
\begin{figure*}[t]
    \centering
    \includegraphics[width=\textwidth]{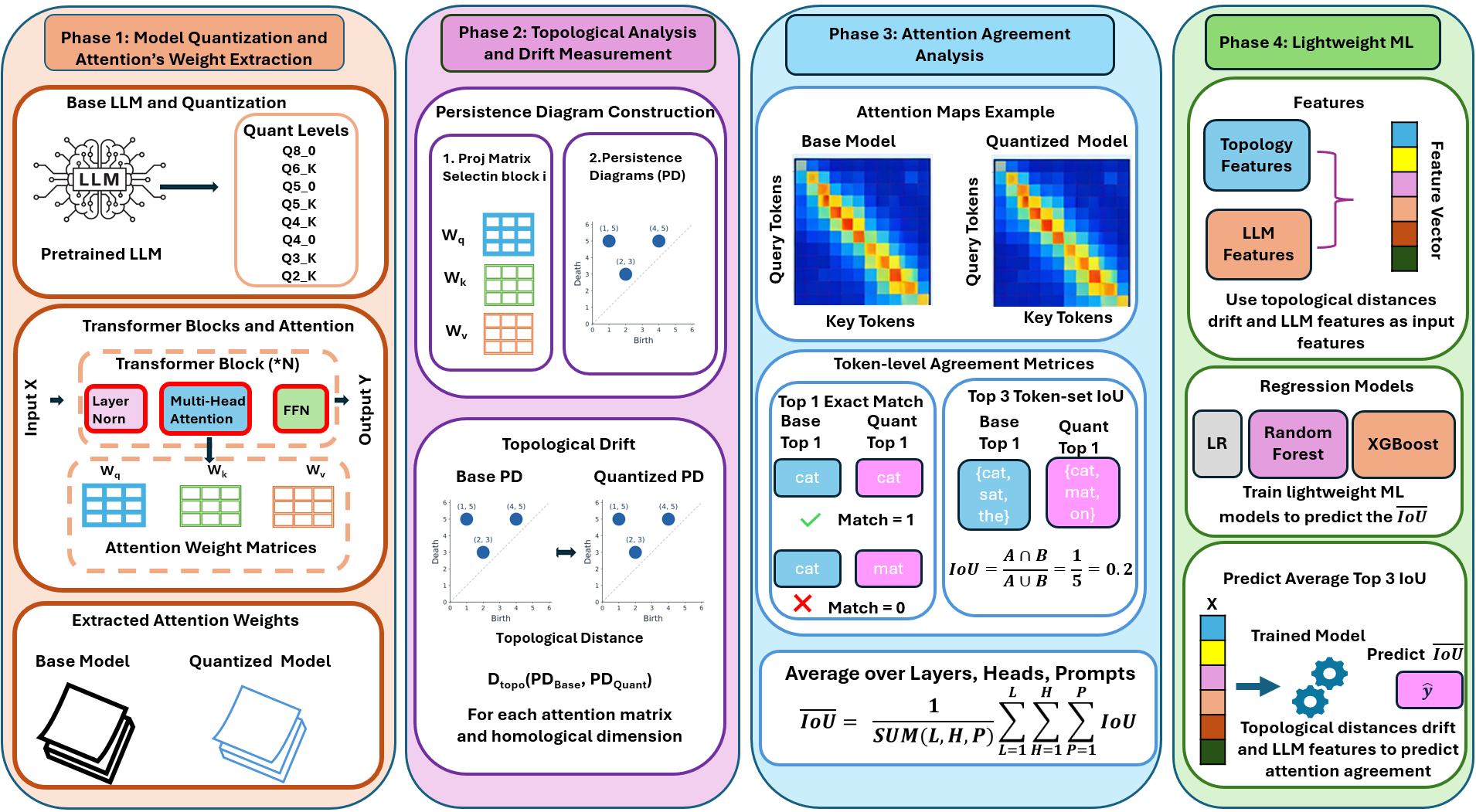}
    \caption{Overview of the TopoQuant framework. The framework consists of four stages: model quantization and weight extraction, topological drift measurement, attention agreement analysis, and lightweight machine learning prediction. Topological features extracted from persistence diagrams are combined with model characteristics to predict attention consistency between base and quantized models.}
    \label{fig:topoquant_framework}
\end{figure*}

TopoQuant framework, illustrated in Figure \ref{fig:topoquant_framework}, consists of four main stages: (1) model quantization and attention-weight extraction, (2) topological analysis and drift measurement, (3) attention agreement analysis, and (4) lightweight machine learning based prediction. At a high level, TopoQuant extracts self-attention weight matrices (Q, K, V) from base and quantized LLMs, constructs persistence diagrams using persistent homology to compute multi-metric topological drift across transformer blocks, and measures functional attention consistency using Top-1 Exact Match and Top-3 IoU. These topological features are combined with model characteristics and passed to lightweight machine learning regressor to predict downstream attention behavioral shifts. 

\subsection{ Model Quantization and Attention Weight Extraction}
\label{subsec: Model_Quantization_and_Attention_Weight_Extraction}

The first stage is to construct multiple quantized variants of a pre-trained LLM at different quantization settings. The full-precision model is called the base model. For a given transformer block $\ell$, the attention projections are characterized by the query, key, and value weight matrices, $W_Q$, $W_K$, $W_V$. The weight matrices are extracted from both the base ($M_O$) and each quantized variant ($M_q$). We extract weights across all blocks to systematically compare quantization changes at different depths. For each transformer block $i$ $\in [N] := \{1, \dots, N\}$ and attention projection $L \in \{ Q, K, V \}$, the framework obtains a pair of matrices $W_{L,\text{base}}^{(i)} \quad \text{and} \quad W_{L,\text{quant}}^{(i)}$, which serve as the input to the topological analysis. 

\subsection{Topological Analysis and Drift Measurement}
\label{subsec:  Drift_Measurement}


TopoQuant characterizes the quantization-induced structural drift by measuring the distances between persistence diagrams generated from the base and quantized attention weight matrices.
We constructed an appropriate representation and filtration to generate the corresponding persistence diagram (PD). Let $\text{PD}_{\text{base}}^{(i, L)} \quad \text{and} \quad \text{PD}_{\text{quant}}^{(i, L)})$ denote the persistence diagrams obtained from the base and quantized models, respectively, for projection $L$ in transformer block $i$. Then, the topological differences between the two models are measured using $D_{\text{topo}}^{(i, L)} = d_{\text{topo}}\!\left(\text{PD}_{\text{base}}^{(i, L)}, \text{PD}_{\text{quant}}^{(i, L)}\right)$, where the distance represents the topological drift by quantization. We repeated this step across the transformer blocks, attention projections, quantization settings, and selected topological distances.

The framework measures the structural drift between the persistence diagrams of the base and quantized models using a diverse set of topological distance metrics. These measures capture the distinct geometric and statistical properties of the parameter manifolds: the Betti distance and Persistence Landscape measure the discretized $L_2$ norm between Betti curves and landscape functions, respectively; Persistence Entropy captures the overall structural complexity by computing the absolute difference between topological entropy values; the Persistence Fisher Distance (PFD) computes the Fisher–Rao distance between kernel density representations of the topological points; and Persistence Images (PI) convert diagrams into weighted two-dimensional Gaussian functions to calculate their Euclidean distance. Finally, the Sliced Wasserstein Distance (SWD) and Tree-Sliced Wasserstein Distance (TSWD) project persistence points onto one-dimensional lines and dyadic tree structures, respectively, to aggregate projection differences across multiple directions.

\subsection{Attention Agreement Analysis}
\label{subsec: Attention_Agreement_Analysis}

The third stage evaluates whether the structural changes identified in Phase 2 correspond to changes in the model's functional attention behavior. Therefore, we processed the same evaluation prompts independently with the base and quantized models and collected their attention maps for comparison. We considered two complementary measures of attention agreement. 

\textbf{Top-1 exact-match agreement.} For each query token, the key token receiving the highest attention value is identified independently for the base and quantized models. $t_{\text{base}}^\star = \arg\max_j A_{\text{base}}(i,j) \quad$ \text{and} $\quad t_{\text{quant}}^\star = \arg\max_j A_{\text{quant}}(i,j)$, where A(i,j) denotes the attention value from the query token to the key token. The Top-1 agreement indicator is defined as 

$M_i = \begin{cases} 
1 & \text{if } t_{\text{base}}^\star = t_{\text{quant}}^\star, \\ 
0 & \text{otherwise.} 
\end{cases}$

\textbf{Top-3 attention agreement.} To provide a less restrictive measure of attention preservation, the three most strongly attended tokens are extracted from the base and quantized variant. Let $S_{\text{base}}(i) \quad$ and $\quad S_{\text{quant}}(i)$ represent the corresponding Top-3 token sets for query token in block i. Their agreement is quantified using intersection over union: 
$\text{IoU}_i = \frac{\left| S_{\text{base}}(i) \cap S_{\text{quant}}(i) \right|}{\left| S_{\text{base}}(i) \cup S_{\text{quant}}(i) \right|}$. The resulting agreement scores are then aggregated across the relevant query tokens, attention heads, blocks, and evaluation prompts to produce the average IoU agreement. 

\subsection{Lightweight Machine Learning Prediction}
\label{subsec:Lightweight_Machine_Learning_Prediction}
The final phase investigates whether topological drift measured from the attention weights can predict the average IoU agreement between the base and quantized variants for each block. Predicting attention agreement is valuable because it provides an early estimate of quantization-induced behavioral degradation without requiring expensive downstream evaluation. If attention agreement can be accurately predicted from structural features such as topological drift, practitioners can identify problematic quantization settings and select safe compression levels before performing extensive benchmarking.

We organize the topological distance drift from Step 2 and the LLM features (family, number of parameters, overall quantization level, quantization level of each projection, and block category [beginning, middle, late]) into feature vectors that capture the structural changes associated with each quantization configuration. We categorize LLM blocks into \textit{Early}, \textit{Middle}, and \textit{Late} based on~\cite{TamoCRPHW26}, which reveals a three-phase internal structure: early blocks align inputs into a shared semantic space, middle blocks perform task reasoning, and late blocks drive language-specific generation. A generic feature vector can be expressed as $\mathbf{x} = \Big[ f_{\mathrm{family}}, f_{\mathrm{params}}, f_{\mathrm{quant}}, f_{\mathrm{Q\_Quant}}, f_{\mathrm{K\_Quant}}, f_{\mathrm{V\_Quant}}, \allowbreak f_{\mathrm{layer\_category}}, D_{\text{topo}}^{(1, Q)}, D_{\text{topo}}^{(1, K)}, D_{\text{topo}}^{(1, V)}, \dots, D_{\text{topo}}^{(L, V)} \Big]$. These feature vectors are paired with $\overline{\mathrm{IoU}}$ obtained in phase 3, where the Top-3 $\overline{\mathrm{IoU}}$ serves as the prediction target. The framework constructs feature configurations based on individual attention projection matrices ((Q, K, V), pairwise combinations (KQ, KV, and QV), and their joint representation (KQV) to study the effect of single and group projection in predicting the Top-3 $\overline{\mathrm{IoU}}$. For each feature configuration, we train and evaluate three lightweight machine learning algorithms (LinearRegression, RandomForest, and XGBoost). We compare the models' predictive performance based on prediction error, and select the model with the lowest error as the best-performing configuration.

\section{Evaluations}
\label{sec:Evaluations}

To evaluate TopoQuant, we addressed the following research questions:

\begin{itemize}
    \item \textbf{RQ1:} How do the topological structure of LLM weights and attention agreement change across quantization levels?
    \item \textbf{RQ2:} Can TDA reliably determine the optimal stopping point in LLM quantization without fully retraining or re-evaluating the model?
    \item \textbf{RQ3:} Can TDA detect the behavioral shift?
    \item \textbf{RQ4:} RQ4: Which attention projection matrices are most informative for quantization analysis?
    \item \textbf{RQ5:} What are the ablation study results for various TopoQuant components and hyperparameters?
\end{itemize}

\subsection{Implementation.}
\label{subsec:Implementation.}

\textbf{Models:} We quantize the following four models: \texttt{Llama-3.2-3B}, \texttt{Vicuna-7B-v1.5}, \texttt{Mistral-7B-v0.1}, and \texttt{Llama-3.1-8B} into legacy quantization methods (\texttt{Q8\_0}, \texttt{Q5\_0}, \texttt{Q4\_0}) and K-quantization methods(\texttt{Q6\_K}, \texttt{Q5\_K}, \texttt{Q4\_K}, \texttt{Q3\_K}, \texttt{Q2\_K}) using llama.cpp \cite{gerganov2024llamacpp}. We extract the attention matrices (Q, K, V) weights of each block of the base models and its quantized variants after dequantization. Notably, K-quantization methods allocate higher precision to the value ($V$) projection than to the other weights in the quantized configurations considered. Specifically, the $V$ projection uses \texttt{Q6\_K} in both \texttt{Q5\_K} and \texttt{Q4\_K}, \texttt{Q5\_K} in \texttt{Q3\_K}, and \texttt{Q4\_K} in \texttt{Q2\_K}.  Downstream task evaluations (accuracy and perplexity) were conducted on an NVIDIA Tesla H100 80GB PCIe GPU, while TopoQuant calculations were performed on a workstation equipped with an Intel Core i7-14700F CPU (2.10 GHz), 32 GB RAM, and an NVIDIA GeForce RTX 3090 GPU.

\textbf{Persistence Diagram:} We convert each attention weight matrix $\mathbf{X}$ into an $N \times N$ pairwise cosine distance matrix $\mathbf{D}$. The distance matrix is then provided to Ripser~\cite{Bauer2021Ripser} with a filtration threshold of 0.8. As the filtration radius $r$ increases, Ripser records the birth and death scales $(r_{\text{birth}}, r_{\text{death}})$ of connected components ($H_0$). The resulting persistence diagram captures the evolution of connectivity within the attention weight space and is visualized using Persim~\cite{persim2026} and serialized into \texttt{.pkl} files for subsequent analysis.

\textbf{Topological Drift.} We compute the topological distance metrics introduced in Section~\ref{subsec: Drift_Measurement} between the persistence diagrams of the base and quantized models across layers and attention projections (Q, K, V). We save the resulting distances, runtime, and layer metadata for benchmarking.

\textbf{Datasets:} We use zero-shot multiple-choice benchmarks including ARC (AI2 Reasoning Challenge)~\cite{clark2018think}, HellaSwag~\cite{zellers2019hellaswag}, Winogrande~\cite{sakaguchi2019winogrande}. ARC (AI2 Reasoning Challenge)~\cite{clark2018think} evaluates grade-school science question answering. HellaSwag~\cite{zellers2019hellaswag} evaluates commonsense reasoning through context completion. WinoGrande~\cite{sakaguchi2019winogrande} evaluates commonsense reasoning and pronoun resolution. 

\textbf{Attention Agreement.} We evaluate base and quantized models on 500 samples from each dataset and extract block-wise self-attention scores. For each prompt, we analyze the attention distribution of the final query token while masking self-attention and special tokens. We identify the top-3 attended context tokens and compare quantized and base models using Top-1 Exact Match and Top-3 Intersection-over-Union (IoU) after aligning records by [model\_name, block]. Item-level, block-level, and global summaries are then generated.

We use top-3 attention agreement for three reasons. First, prior studies show that only a small subset of tokens typically receives most of the attention mass in decoder-only LLMs \cite{Zhang00CZC0TRBW23,XiaoTCHL24}. Second, top-1 matching is overly sensitive to minor ranking fluctuations that do not necessarily reflect meaningful behavioral changes. Third, larger sets (top-5 +) introduce additional noise. Therefore, top-3 provides a practical balance between stability and sensitivity when detecting quantization-induced attention drift.

\textbf{Lightweight Machine Learning Prediction.} We use the topological drift features and model metadata described in Section~\ref{subsec:Lightweight_Machine_Learning_Prediction} to predict $\overline{\mathrm{IoU}}$. The feature configurations include individual projections (Q, K, V), pairwise combinations (KQ, KV, QV), and the combined representation (KQV). Linear regression, random forest, and XGBoost were evaluated using 5-fold cross-validation.

\begin{figure*}[t]
    \centering
    \includegraphics[width=\textwidth]{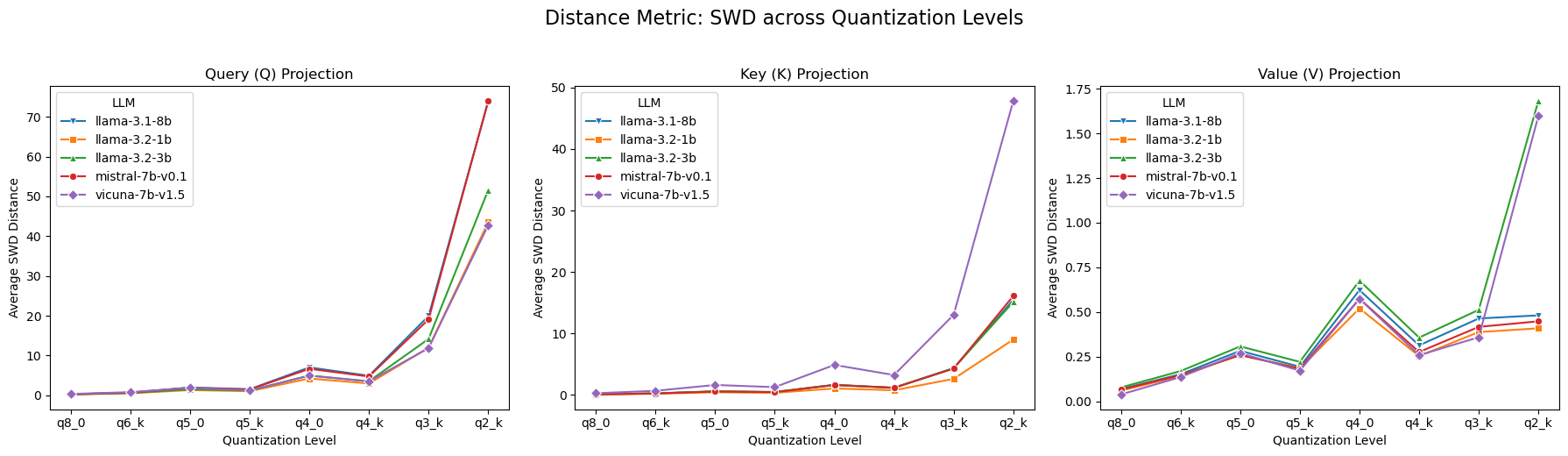}
    \caption{Sliced Wasserstein Distance (SWD) drift across quantization levels for Query ($\mathbf{Q}$), Key ($\mathbf{K}$), and Value ($\mathbf{V}$) projection matrices.}
    \label{fig:swd_distances}
\end{figure*}

\begin{table*}[t]
\centering
\small
\setlength{\tabcolsep}{3.5pt}
\renewcommand{\arraystretch}{1.05}
\caption{\textbf{Attention Agreement Metrics across Quantization Configurations.} Evaluation of Top-3 Intersection over Union (IoU) and Top-1 Agreement for ARC, HellaSwag, and Winogrande datasets across four LLM architectures.}
\label{tab:combined_attention_agreement}
\begin{tabular}{ll ccc ccc ccc ccc}
\toprule
 & & \multicolumn{3}{c}{\textbf{Llama-3B}} & \multicolumn{3}{c}{\textbf{Vicuna-7B}} & \multicolumn{3}{c}{\textbf{Mistral-7B}} & \multicolumn{3}{c}{\textbf{Llama-8B}} \\
\cmidrule(r){3-5} \cmidrule(lr){6-8} \cmidrule(lr){9-11} \cmidrule(l){12-14}
\textbf{Metric} & \textbf{Quant} & \textbf{Arc} & \textbf{Hell.} & \textbf{Wino.} & \textbf{Arc} & \textbf{Hell.} & \textbf{Wino.} & \textbf{Arc} & \textbf{Hell.} & \textbf{Wino.} & \textbf{Arc} & \textbf{Hell.} & \textbf{Wino.} \\
\midrule

\multirow{8}{*}{\textbf{\shortstack[l]{Top-1\\Agreement}}}
& Q8\_0 & \best{0.74} & \best{0.88} & \best{0.68} & \best{0.99} & \best{1.00} & \best{1.00} & \best{0.99} & \best{1.00} & \best{0.99} & \best{0.75} & \best{0.87} & \best{0.71} \\ 
& Q6\_K & \best{0.74} & \best{0.88} & 0.67 & 0.97 & \best{1.00} & 0.99 & 0.97 & 0.98 & 0.96 & \best{0.75} & \best{0.87} & \best{0.71} \\ 
& Q5\_0 & 0.73 & \best{0.88} & \best{0.68} & 0.94 & 0.99 & 0.98 & 0.93 & 0.96 & 0.92 & 0.73 & 0.86 & 0.68 \\ & Q5\_K & \best{0.74} & \best{0.88} & 0.67 & 0.96 & 0.99 & 0.99 & 0.96 & 0.97 & 0.95 & \best{0.75} & 0.86 & \best{0.71} \\ 
& Q4\_0 & 0.72 & 0.87 & 0.67 & 0.88 & 0.99 & 0.95 & 0.89 & 0.93 & 0.83 & \worst{0.60} & 0.85 & 0.68 \\ 
& Q4\_K & 0.73 & 0.87 & 0.67 & 0.93 & 0.99 & 0.98 & 0.92 & 0.95 & 0.90 & 0.72 & 0.86 & 0.70 \\ 
& Q3\_K & 0.70 & 0.85 & 0.64 & 0.87 & 0.98 & 0.95 & 0.86 & 0.91 & 0.83 & 0.71 & 0.84 & 0.66 \\ 
& Q2\_K & \worst{0.61} & \worst{0.78} & \worst{0.56} & \worst{0.74} & \worst{0.98} & \worst{0.86} & \worst{0.66} & \worst{0.84} & \worst{0.71} & 0.61 & \worst{0.77} & \worst{0.60} \\ 
\midrule \multirow{8}{*}{\textbf{\shortstack[l]{Top-3\\IoU}}} 
& Q8\_0 & \best{0.75} & \best{0.83} & 0.70 & \best{1.00} & \best{0.99} & \best{0.99} & \best{0.99} & \best{0.99} & \best{0.99} & \best{0.74} & \best{0.83} & \best{0.73} \\ 
& Q6\_K & \best{0.75} & \best{0.83} & 0.70 & 0.99 & 0.98 & 0.97 & 0.97 & 0.98 & 0.96 & 0.73 & \best{0.83} & \best{0.73} \\ 
& Q5\_0 & 0.74 & 0.82 & \best{0.71} & 0.97 & 0.96 & 0.93 & 0.94 & 0.95 & 0.93 & 0.72 & \best{0.83} & 0.72 \\ 
& Q5\_K & \best{0.75} & \best{0.83} & \best{0.71} & 0.98 & 0.97 & 0.96 & 0.96 & 0.97 & 0.95 & 0.73 & \best{0.83} & \best{0.73} \\ 
& Q4\_0 & 0.74 & 0.81 & 0.69 & 0.94 & 0.92 & 0.86 & 0.89 & 0.92 & 0.84 & 0.70 & 0.82 & 0.71 \\ 
& Q4\_K & 0.74 & 0.82 & 0.70 & 0.97 & 0.95 & 0.93 & 0.93 & 0.94 & 0.91 & 0.72 & 0.82 & 0.72 \\ 
& Q3\_K & 0.69 & 0.80 & 0.67 & 0.93 & 0.91 & 0.89 & 0.87 & 0.90 & 0.84 & 0.70 & 0.81 & 0.70 \\ 
& Q2\_K & \worst{0.61} & \worst{0.73} & \worst{0.59} & \worst{0.86} & \worst{0.83} & \worst{0.73} & \worst{0.66} & \worst{0.82} & \worst{0.72} & \worst{0.58} & \worst{0.75} & \worst{0.63} \\
\bottomrule
\end{tabular}
\end{table*}

\textbf{RQ1: topological structure and attention agreement change across quantization levels:} Both topological structure and attention agreement remain largely stable under moderate quantization before degrading sharply at sub-4-bit precision. As shown in Figure~\ref{fig:swd_distances} and Table~\ref{tab:combined_attention_agreement}, low structural drift corresponds closely to high attention agreement. At higher precisions ($\text{Q8\_0}$ to $\text{Q5\_K}$), SWD drift remains near zero while Top-1 Agreement and Top-3 IoU typically exceed 0.98. In contrast, aggressive quantization ($\text{Q3\_K}$ and $\text{Q2\_K}$) produces large increases in topological drift accompanied by substantial declines in attention agreement, with Top-1 Agreement dropping to 0.61 and Top-3 IoU falling to 0.58 in some configurations. The degradation is concentrated in the Q and K projections, which exhibit substantially larger topological distortions than the V projection. The lower drift observed in the V projection may be partly influenced by the K-quantization schemes used in this study, which allocate higher precision to V than to other attention weights. Furthermore, mid-sized models such as Vicuna-7B and Mistral-7B preserve both structural and behavioral consistency more effectively than smaller models such as Llama-3B across all quantization levels.

\begin{figure*}[t]
    \centering
    \includegraphics[width=\textwidth]{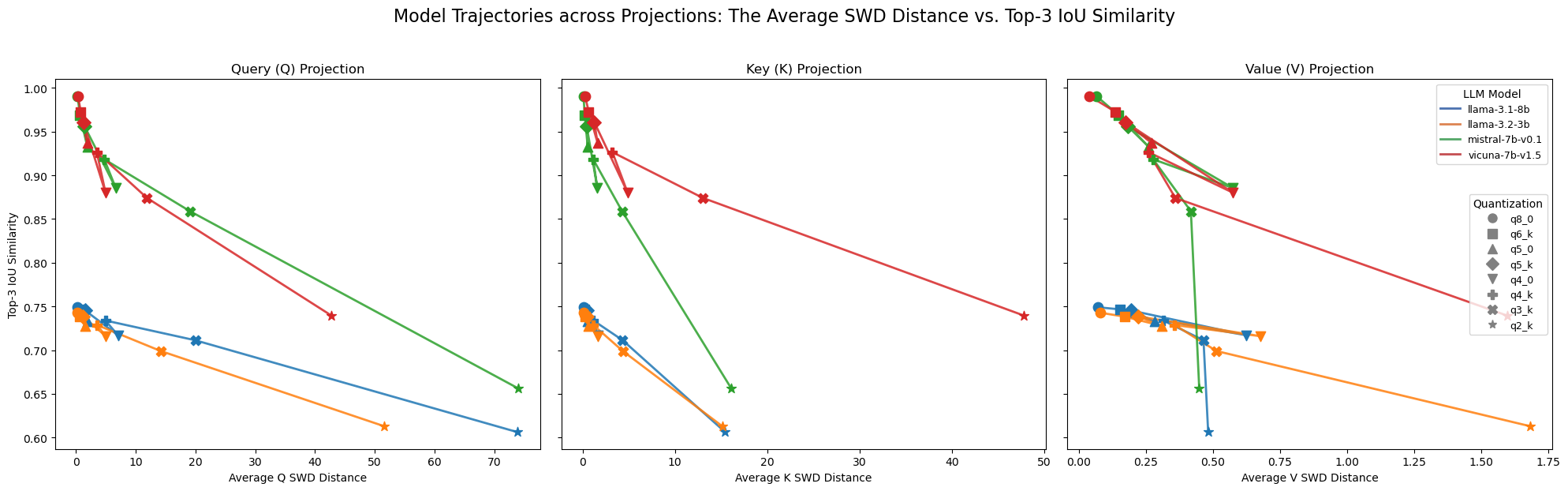}
    \caption{\textbf{Quantization degradation trajectories across attention projection matrices.} Each curve tracks an LLM model across eight quantization level (ranging from high-precision $\mathbf{q8\_0}$ to extreme low-bit $\mathbf{q2\_k}$, indicated by marker shape). The $x$-axis shows the average Sliced Wasserstein Distance (SWD) structural drift, while the $y$-axis displays downstream behavioral preservation measured via Top-3 IoU Similarity on the ARC benchmark. The exponential divergence in Query ($Q$) and Key ($K$) projections highlights their structural vulnerability, validating SWD as an efficient proxy for tracking performance cliffs without full benchmark re-evaluation.}
    \label{fig:swd_trajectories}
\end{figure*}


\textbf{RQ2: TDA as a reliable metric:} The results suggest that TDA can identify practical quantization stopping points without exhaustive downstream evaluation. As shown in Figure~\ref{fig:swd_trajectories}, SWD exhibits a consistent inverse relationship with Top-3 IoU, where increasing structural drift is accompanied by declining attention agreement. Quantization levels from $\text{Q8\_0}$ to $\text{Q5\_K}$ remain clustered in a low-drift, high-agreement region, whereas $\text{Q3\_K}$ and $\text{Q2\_K}$ move rapidly toward high-drift, low-agreement regions. This clear transition indicates that topological drift can serve as an effective indicator of quantization-induced degradation and help identify safe compression limits. Query ($Q$) and Key ($K$) projections exhibit the strongest correspondence between structural drift and attention degradation, while the Value ($V$) projection shows consistently lower drift, potentially influenced by the higher precision assigned to $V$ in the K-quantization schemes used in this study. Similar trends were observed across the other evaluated topological distance metrics, including Betti distance, persistence entropy, persistence landscapes, persistence Fisher distance (PFD), persistence images, and tree-sliced Wasserstein distance (TSWD). These findings suggest that TDA provides a lightweight alternative for monitoring quantization quality and identifying performance cliffs without requiring extensive downstream benchmarking.

\begin{table*}[t]
\centering
\caption{\textbf{Ablation Study of Feature Representations on Attention Agreement ($R^2$ for Top-3 IoU) across Benchmarks.} Comparison across LLM features only, TDA features only, and Combined ($\text{LLM} + \text{TDA}$) feature configurations. Top values per column are highlighted in \colorbox{green!25}{green} and lowest values in \colorbox{red!25}{red}.}
\label{tab:ablation_r2_results}
\resizebox{\textwidth}{!}{%
\begin{tabular}{ll c ccc c ccc c ccc}
\toprule
& & & \multicolumn{3}{c}{\textbf{Arc}} & & \multicolumn{3}{c}{\textbf{Hell.}} & & \multicolumn{3}{c}{\textbf{Wino.}} \\
\cmidrule{4-6} \cmidrule{8-10} \cmidrule{12-14}
\textbf{Feature Matrix} & \textbf{Algorithm} & & \textbf{LLM} & \textbf{TDA} & \textbf{Comb.} & & \textbf{LLM} & \textbf{TDA} & \textbf{Comb.} & & \textbf{LLM} & \textbf{TDA} & \textbf{Comb.} \\
\midrule

\multirow{3}{*}{\textbf{Query ($\mathbf{Q}$)}}
& LinearRegression & & \colorbox{green!25}{\textbf{0.68}} & 0.26 & 0.76 & & 0.53 & 0.27 & 0.83 & & \colorbox{green!25}{\textbf{0.75}} & 0.22 & 0.81 \\
& XGBoost          & & \colorbox{red!25}{0.49} & 0.21 & 0.66 & & \colorbox{red!25}{0.47} & 0.11 & 0.51 & & \colorbox{red!25}{0.56} & 0.20 & 0.66 \\
& RandomForest     & & 0.67 & \colorbox{green!25}{\textbf{0.33}} & \colorbox{green!25}{\textbf{0.83}} & & \colorbox{green!25}{\textbf{0.72}} & 0.35 & \colorbox{green!25}{\textbf{0.85}} & & 0.74 & 0.31 & \colorbox{green!25}{\textbf{0.83}} \\
\midrule

\multirow{3}{*}{\textbf{Key ($\mathbf{K}$)}}
& LinearRegression & & \colorbox{green!25}{\textbf{0.68}} & 0.21 & 0.73 & & 0.53 & 0.22 & 0.82 & & \colorbox{green!25}{\textbf{0.75}} & 0.19 & 0.82 \\
& XGBoost          & & \colorbox{red!25}{0.49} & 0.21 & 0.65 & & \colorbox{red!25}{0.47} & 0.09 & 0.50 & & \colorbox{red!25}{0.56} & 0.19 & 0.65 \\
& RandomForest     & & 0.67 & \colorbox{green!25}{\textbf{0.33}} & 0.82 & & \colorbox{green!25}{\textbf{0.72}} & 0.39 & 0.84 & & 0.74 & 0.35 & 0.82 \\
\midrule

\multirow{3}{*}{\textbf{Value ($\mathbf{V}$)}}
& LinearRegression & & \colorbox{green!25}{\textbf{0.68}} & \colorbox{red!25}{--} & 0.75 & & 0.53 & \colorbox{red!25}{--} & 0.83 & & \colorbox{green!25}{\textbf{0.75}} & \colorbox{red!25}{--} & \colorbox{red!25}{0.15} \\
& XGBoost          & & \colorbox{red!25}{0.49} & 0.21 & \colorbox{red!25}{0.60} & & \colorbox{red!25}{0.47} & \colorbox{red!25}{0.03} & \colorbox{red!25}{0.47} & & \colorbox{red!25}{0.56} & \colorbox{red!25}{0.18} & 0.60 \\
& RandomForest     & & 0.67 & \colorbox{green!25}{\textbf{0.33}} & 0.75 & & \colorbox{green!25}{\textbf{0.72}} & 0.11 & 0.74 & & 0.74 & 0.29 & 0.75 \\
\midrule

\multirow{3}{*}{\textbf{Pairwise ($\mathbf{KQ}$)}}
& LinearRegression & & \colorbox{green!25}{\textbf{0.68}} & 0.28 & 0.76 & & 0.53 & 0.29 & 0.83 & & \colorbox{green!25}{\textbf{0.75}} & 0.26 & 0.82 \\
& XGBoost          & & \colorbox{red!25}{0.49} & 0.21 & 0.67 & & \colorbox{red!25}{0.47} & 0.11 & 0.51 & & \colorbox{red!25}{0.56} & 0.21 & 0.67 \\
& RandomForest     & & 0.67 & \colorbox{green!25}{\textbf{0.33}} & 0.82 & & \colorbox{green!25}{\textbf{0.72}} & 0.39 & 0.84 & & 0.74 & \colorbox{green!25}{\textbf{0.37}} & 0.82 \\
\midrule

\multirow{3}{*}{\textbf{Pairwise ($\mathbf{KV}$)}}
& LinearRegression & & \colorbox{green!25}{\textbf{0.68}} & \colorbox{red!25}{--} & 0.73 & & 0.53 & \colorbox{red!25}{--} & 0.82 & & \colorbox{green!25}{\textbf{0.75}} & \colorbox{red!25}{--} & 0.72 \\
& XGBoost          & & \colorbox{red!25}{0.49} & 0.21 & 0.65 & & \colorbox{red!25}{0.47} & 0.09 & 0.50 & & \colorbox{red!25}{0.56} & 0.22 & 0.65 \\
& RandomForest     & & 0.67 & \colorbox{green!25}{\textbf{0.33}} & 0.82 & & \colorbox{green!25}{\textbf{0.72}} & 0.38 & 0.84 & & 0.74 & 0.33 & 0.82 \\
\midrule

\multirow{3}{*}{\textbf{Pairwise ($\mathbf{QV}$)}}
& LinearRegression & & \colorbox{green!25}{\textbf{0.68}} & \colorbox{red!25}{--} & 0.76 & & 0.53 & \colorbox{red!25}{--} & 0.83 & & \colorbox{green!25}{\textbf{0.75}} & \colorbox{red!25}{--} & 0.69 \\
& XGBoost          & & \colorbox{red!25}{0.49} & 0.21 & 0.67 & & \colorbox{red!25}{0.47} & 0.11 & 0.51 & & \colorbox{red!25}{0.56} & 0.25 & 0.67 \\
& RandomForest     & & 0.67 & \colorbox{green!25}{\textbf{0.33}} & 0.82 & & \colorbox{green!25}{\textbf{0.72}} & 0.36 & 0.84 & & 0.74 & 0.34 & 0.82 \\
\midrule

\multirow{3}{*}{\textbf{Joint ($\mathbf{KQV}$)}}
& LinearRegression & & \colorbox{green!25}{\textbf{0.68}} & 0.31 & 0.75 & & 0.53 & \colorbox{red!25}{--} & 0.72 & & \colorbox{green!25}{\textbf{0.75}} & \colorbox{red!25}{--} & 0.73 \\
& XGBoost          & & \colorbox{red!25}{0.49} & 0.21 & 0.65 & & \colorbox{red!25}{0.47} & 0.11 & 0.51 & & \colorbox{red!25}{0.56} & 0.24 & 0.65 \\
& RandomForest     & & 0.67 & 0.31 & 0.82 & & \colorbox{green!25}{\textbf{0.72}} & \colorbox{green!25}{\textbf{0.41}} & \colorbox{green!25}{\textbf{0.85}} & & 0.74 & 0.32 & 0.82 \\
\bottomrule
\end{tabular}
}
\end{table*}

\textbf{RQ3: TDA and the behavioral shift} TDA integrated with LLM representations (the combined feature set) effectively predicts the behavioral shift (measured by $\overline{\mathrm{IoU}}$) across benchmark datasets. Based on the performance evaluation in Table~\ref{tab:ablation_r2_results}: (1) The combined feature representation achieves high predictive performance across all benchmark datasets, consistently reaching $R^2$ scores of up to $0.85$ on HellaSwag, $0.83$ on Arc, and $0.83$ on Winogrande. (2) Both Random Forest and Linear Regression demonstrate superior predictive capabilities within the combined feature space compared to XGBoost. Random Forest consistently delivers robust scores across all matrix configurations ($R^2 \approx 0.74$--$0.85$), while Linear Regression achieves its strongest performance on HellaSwag and Winogrande ($R^2 \approx 0.81$--$0.83$). (3) High predictive power is maintained across individual matrices (Q, K, V), pairwise interactions (KQ, KV, QV), and the joint configuration (KQV), confirming that TopoQuant captures a reliable and stable signal for predicting attention agreement shifts.

%
\textbf{RQ4:RQ4: attention projection matrices most informative for quantization analysis:} Based on the combined feature evaluation ($\text{LLM} + \text{TDA}$) in Table~\ref{tab:ablation_r2_results}, the Q and K matrices, along with their pairwise interaction (KQ), emerge as the most informative indicators of quantization-induced attention degradation. Features derived from Q and K consistently achieve the highest predictive performance across all benchmarks. On HellaSwag, both Q and K reach peak $R^2$ scores of $0.85$ (Random Forest) and $0.83$ (Linear Regression), outperforming or matching all other configurations. Similarly, the KQ combination achieves strong predictive accuracy across ARC ($R^2 = 0.76$--$0.82$), HellaSwag ($R^2 = 0.84$), and Winogrande ($R^2 = 0.82$), indicating that structural changes within the attention routing mechanism are closely associated with attention agreement degradation. In contrast, configurations relying solely on the V projection generally exhibit lower predictive performance. For example, Linear Regression achieves an $R^2$ of only $0.15$ on Winogrande using V alone, compared with $0.81$ and $0.82$ for Q and K, respectively. Although Random Forest improves V-based performance ($R^2 = 0.74$--$0.75$), it remains below the best-performing Q, K, and KQ based configurations. Combining all three matrices (KQV) maintains strong performance ($R^2$ up to $0.85$), but does not meaningfully outperform Q, K, or KQ alone. These results suggest that structural changes within the Q and K projections are more strongly associated with attention degradation than those observed in the V projection, making them particularly useful for monitoring quantization-induced behavioral drift.

\textbf{RQ5: ablations on feature representations and TopoQuant components} To isolate the individual predictive contributions of topological representation learning, we conduct a systematic ablation study comparing three distinct feature configurations: (1) \textbf{Model Features Only} (comprising 5 base configuration and layer-category indicators: \texttt{quant}, \texttt{layer\_category}, \texttt{q\_quant}, \texttt{k\_quant}, and \texttt{v\_quant}), (2) \textbf{TDA Features Only} (comprising 21 topological persistence features extracted across the $Q$, $K$, and $V$ weight matrices—specifically Betti curves, entropy, landscapes, persistence images, SWD and TSWD), and (3) \textbf{Combined Features} ($5 + 21 = 26$ total dimensions formed by concatenating the model and TDA feature spaces), as shown in Table~\ref{tab:ablation_r2_results}. The evaluation yields the following key insights regarding feature contributions across architectures and benchmarks: (1) Across all benchmarks and attention matrix configurations, the combined feature space consistently yields the highest predictive accuracy ($R^2$),  outperforming both individual feature sets. On HellaSwag, the combined features raise $R^2$ to $0.85$ (Random Forest across Q, K, and KQV), compared with $0.72$ for LLM features only and $0.41$ for standalone TDA features, confirming that topological metrics provide complementary signals that improve baseline model performance. (2) While LLM Features Only offered a moderate predictive baseline ($R^2 \approx 0.67$--$0.75$ on Arc and Winogrande using Random Forest/Linear Regression), they failed to capture subtle behavioral shifts on complex benchmarks. Conversely, TDA Features Only yield modest standalone performance ($R^2 \approx 0.09$--$0.41$), and frequently struggle or produce negative/near-zero fits when evaluated independently on the V matrix under linear assumptions. (3) Non-linear ensemble methods (Random Forest) excelled at leveraging the combined feature space, maintaining stable $R^2$ scores ($0.74$--$0.85$) across all matrix views (Q, K, V, KQ, KV, QV, KQV). Linear Regression demonstrates strong performance when combining features on HellaSwag ($R^2 = 0.82$--$0.83$ for Q, K, KQ), but exhibits instability on V-dependent features alone, underscoring the importance of non-linear modeling for capturing complex topological dynamics.

\textbf{TopoQuant overhead vs. downstream runtime:} To demonstrate the runtime efficiency and light overhead of TopoQuant, Table~\ref{tab:topoquant_overhead} compares the runtime (in seconds) required to extract topological metrics—including persistence diagrams, topological distance calculations, and layer-wise attention agreement—against the execution time of the downstream ARC benchmark across various quantization schemes on $\texttt{Llama-3.1-8B}$.

Across all evaluated quantization levels (from 8-bit to 2-bit), the computational overhead introduced by TopoQuant remained a fraction of the total downstream execution duration. For high-to-medium precision variants ($\texttt{Q8\_0}$ through $\texttt{Q4\_K}$), TopoQuant computation requires between 862 and 937 seconds, accounting for approximately $16\%$ to $17\%$ of the downstream evaluation time (which ranges from 4,603 to 4,709 seconds). Even at extreme quantization levels ($\texttt{Q3\_K}$ and $\texttt{Q2\_K}$), where the metric extraction time increases slightly—peaking at 1,250 s for $\texttt{Q3\_K}$—the diagnostic overhead remains within $21.0\%$ $26.5\%$ of the downstream evaluation runtime.

\begin{table}[t]
\centering
\small
\setlength{\tabcolsep}{8pt}
\renewcommand{\arraystretch}{1.05}
\caption{\textbf{TopoQuant overhead vs. downstream runtime:} To demonstrate the effectiveness of our framework, we compare the time required to compute TopoQuant (persistence diagrams, topological distances, and attention agreement on the ARC dataset) against the time required for downstream task execution alone on $\texttt{Llama-3.1-8B}$.}
\begin{tabular}{l cc}
\hline
Quant & TopoQuant & Downstream \\
\hline
Q8\_0 & 862& 4619 \\
Q6\_K & 937& 4647\\
Q5\_0 & 931& 4603 \\
Q5\_K & 937& 4627\\
Q4\_0 & 903& 4709\\
Q4\_K & 914& 4636\\
Q3\_K & 1250& 4708\\
Q2\_K & 1109& 4694\\
\hline
\end{tabular}
\label{tab:topoquant_overhead}
\end{table}

\section{Discussion}
\label{sec:Discussion}

The empirical findings from the TopoQuant evaluation provide several key insights into the structural and functional mechanisms underlying PTQ in LLMs. In this section, we discuss the broader implications for LLM evaluation, interpretability, and quantization design.

\textbf{Topological Drift as an Early Warning Signal:} A central finding of this work is that conventional benchmark accuracy and perplexity can be highly misleading indicators of model integrity under quantization. As shown in \cite{rababah2026illusion} evaluation of the same LLMs we are studying, a quantized model can maintain perplexity and accuracy while simultaneously experiencing shifts in its internal attention allocation patterns—a phenomenon previously termed the \textquote{illusion of preservation}. 

\textbf{TopoQuant} bridges this diagnostic gap by establishing that topological weight drift serves as an early, highly sensitive warning signal. Because topological metrics capture multi-scale structural deformations on the parameter manifold rather than aggregate statistical averages, they detect the onset of weight matrix breakdown well before downstream decision-making collapses. This capability lets practitioners set principled, data-free quantization stopping points (e.g., before the sub-4-bit exponential drift threshold) without running expensive, full-dataset evaluation benchmarks.

\textbf{Asymmetric vulnerability and attention routing bottlenecks:} Weight matrices responsible for attention routing (Q and K) exhibit substantially larger topological drift under aggressive compression than other attention components. This drift closely aligns with the observed reductions in Top-1 Agreement and Top-3 IoU. In contrast, V projections consistently exhibit lower structural drift across models and quantization levels. However, this observation should be interpreted with caution, as the K-quantization schemes used in this study allocate higher precision to the V projection than to other attention weights, which may partially contribute to its increased stability.

\textbf{Computational Efficiency and Runtime Overhead:} 
A critical requirement for structural diagnostics is that metric extraction must remain significantly cheaper than exhaustive downstream task evaluation. As demonstrated in Table~\ref{tab:topoquant_overhead}, TopoQuant introduces low computational overhead relative to full benchmark runs on $\texttt{Llama-3.1-8B}$. These results confirm that TopoQuant provides deep structural and topological diagnostics at a reasonable computational cost, enabling comprehensive quantization analysis without prohibitive runtime penalties relative to standard benchmark evaluations.

\textbf{Predictive Power and Practical Deployment:}
Our lightweight machine learning experiments demonstrate that combining LLM-level features with multi-scale topological drift descriptors ($\text{LLM} + \text{TDA}$) yields highly accurate predictions of downstream attention agreement ($R^2$ scores up to $0.85$). By relying on lightweight topological extraction via persistent homology and non-linear regressors (such as Random Forest), TopoQuant provides an efficient, parameter-space diagnostic pipeline. Model developers can inspect the structural stability of quantized checkpoints without requiring extensive downstream benchmark evaluation, reducing the computational burden associated with evaluating multiple quantization configurations.

\textbf{Limitations and Future Directions}

While TopoQuant provides a robust framework for structural drift analysis, certain limitations remain: (1) Current evaluations focused on self-attention projection matrices (Q, K, V). Extending persistent homology analysis to multi-layer perceptron (MLP) blocks and embedding could provide a more holistic view of full-network topological drift. (2) The current study evaluated standard uniform and K-quantization variants within llama.cpp. Future work will explore applying TopoQuant to non-uniform, outlier-preserving quantization methods such as AWQ and SmoothQuant. (3) Computing persistence diagrams on large weight matrices ($d \times d$) incurs high computational complexity ($O(d^3)$ for exact filtrations). Although sliced variants (SWD, TSWD) mitigate this cost, scaling TDA to models exceeding 70B parameters requires further optimization, such as mini-batch topological sampling or low-rank projection filtrations. (4) The K-quantization formats evaluated in this study allocated different precisions to the Query, Key, and Value projections. Consequently, the lower drift observed in the Value projection may reflect both intrinsic robustness and reduced quantization severity. Future work should investigate projection-specific sensitivity under controlled quantization settings.

\section{Conclusion}
\label{sec:Conclusion}

In this study, we introduced \textbf{TopoQuant}, a topological framework for characterizing and predicting quantization-induced structural and behavioral changes in large language models (LLMs). Moving beyond conventional metrics such as accuracy and perplexity, TopoQuant leverages persistent homology to quantify structural drift in self-attention weight matrices and relate these changes to attention behavior. Our evaluation across multiple LLM families and quantization levels demonstrates a strong relationship between topological drift and attention agreement. Topological drift remains limited at moderate quantization levels but increases sharply under aggressive compression, closely corresponding to declines in Top-1 Agreement and Top-3 IoU. These results indicate that topological representations provide a meaningful view of quantization-induced degradation that is not captured by traditional evaluation metrics alone. We further show that structural changes within the Query and Key projections are more strongly associated with attention degradation than those observed in the Value projection, making them particularly informative for quantization analysis. In addition, combining model-level characteristics with topological features enables lightweight machine learning models to predict attention agreement with $R^2$ scores of up to $0.85$, reducing the need for extensive downstream benchmarking. Overall, TopoQuant provides a practical framework for monitoring quantization quality, identifying safe compression operating regions, and studying the relationship between structural and behavioral changes in quantized LLMs. We hope that this work encourages further exploration of topology-aware methods for efficient evaluation, diagnosis, and optimization of compressed foundation models.
 
\section{Acknowledgment}
\label{sec:Acknowledgment}

This work was partially supported by (a) NSERC (Canada), (b) University of Manitoba, and (c) Red River College Polytechnic. 

\bibliographystyle{unsrt}
\bibliography{ref}



\end{document}